\documentclass{article}
\usepackage[dvips]{graphicx}
\usepackage{graphicx}
\usepackage{arxiv}
\usepackage[utf8]{inputenc} 
\usepackage[T1]{fontenc}    
\usepackage{hyperref}       
\usepackage{breakurl}       
\usepackage{booktabs}       
\usepackage{amsfonts}       
\usepackage{nicefrac}       
\usepackage{microtype}      
\usepackage{bm}
\usepackage{lipsum}
\usepackage{amsmath}
\usepackage{algorithm}
\usepackage{algorithmic}
\usepackage{multirow}
\usepackage{comment}
\title{Accelerating ODE-Based Neural Networks on Low-Cost FPGAs}
\author{
  Hirohisa Watanabe\\
  Keio University\\
  3-14-1 Hiyoshi, Kohoku-ku, Yokohama, Japan\\
  \texttt{watanabe@arc.ics.keio.ac.jp}\\
  \And
  Hiroki Matsutani \\
  Keio University\\
  3-14-1 Hiyoshi, Kohoku-ku, Yokohama, Japan\\
  \texttt{matutani@arc.ics.keio.ac.jp} \\
}

\begin{document}
\maketitle

\begin{abstract}
    ODENet is a deep neural network architecture in which a stacking
    structure of ResNet is implemented with an ordinary differential
    equation (ODE) solver.
    It can reduce the number of parameters and strike a balance between
    accuracy and performance by selecting a proper solver.
    It is also possible to improve the accuracy while keeping the same
    number of parameters on resource-limited edge devices.
    In this paper, using Euler method as an ODE solver, a part of ODENet
    is implemented as a dedicated logic on a low-cost FPGA
    (Field-Programmable Gate Array) board, such as PYNQ-Z2 board.
    As ODENet variants, reduced ODENets (rODENets) each of which
    heavily uses a part of ODENet layers and reduces/eliminates some layers
    differently are proposed and analyzed for low-cost FPGA
    implementation.
    They are evaluated in terms of parameter size, accuracy, execution
    time, and resource utilization on the FPGA.
    The results show that an overall execution time of an rODENet
    variant is improved by up to 2.66 times compared to a pure
    software execution while keeping a comparable accuracy to the
    original ODENet.
\end{abstract}


\section{Introduction}\label{sec:intro}


ResNet \cite{resnet} is a well-known deep neural network (DNN) architecture
with high accuracy.
In addition to conventional forward propagation of DNNs, 
it has shortcut or skip connections that directly add the input of a
layer to the output of the layer.
Since it can mitigate vanishing and exploding gradient problems, we
can stack more layers to improve prediction accuracy.
However, stacking many layers increases the number of parameters of
DNNs; in this case, memory requirement becomes severe
in resource-limited edge devices.

\begin{figure*}[t!]
	\centering
	\begin{tabular}{c}
		\begin{minipage}{0.5\hsize}
			\centering
			\includegraphics[height=60mm]{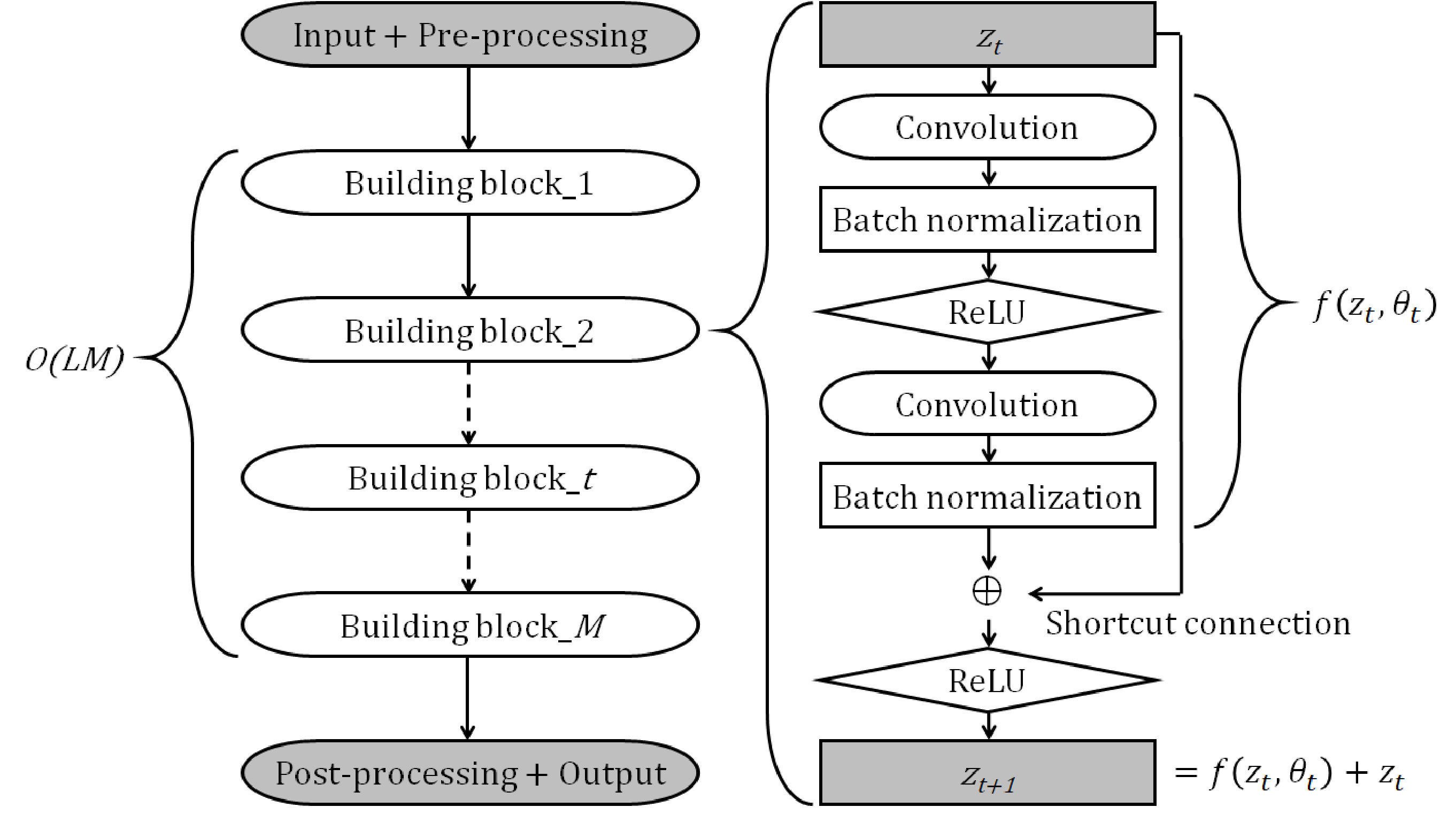}
			\caption{ResNet architecture}
			\label{fig:resnet}
		\end{minipage}
		\hspace{5mm}
		\begin{minipage}{0.48\hsize}
			\centering
			\includegraphics[height=60mm]{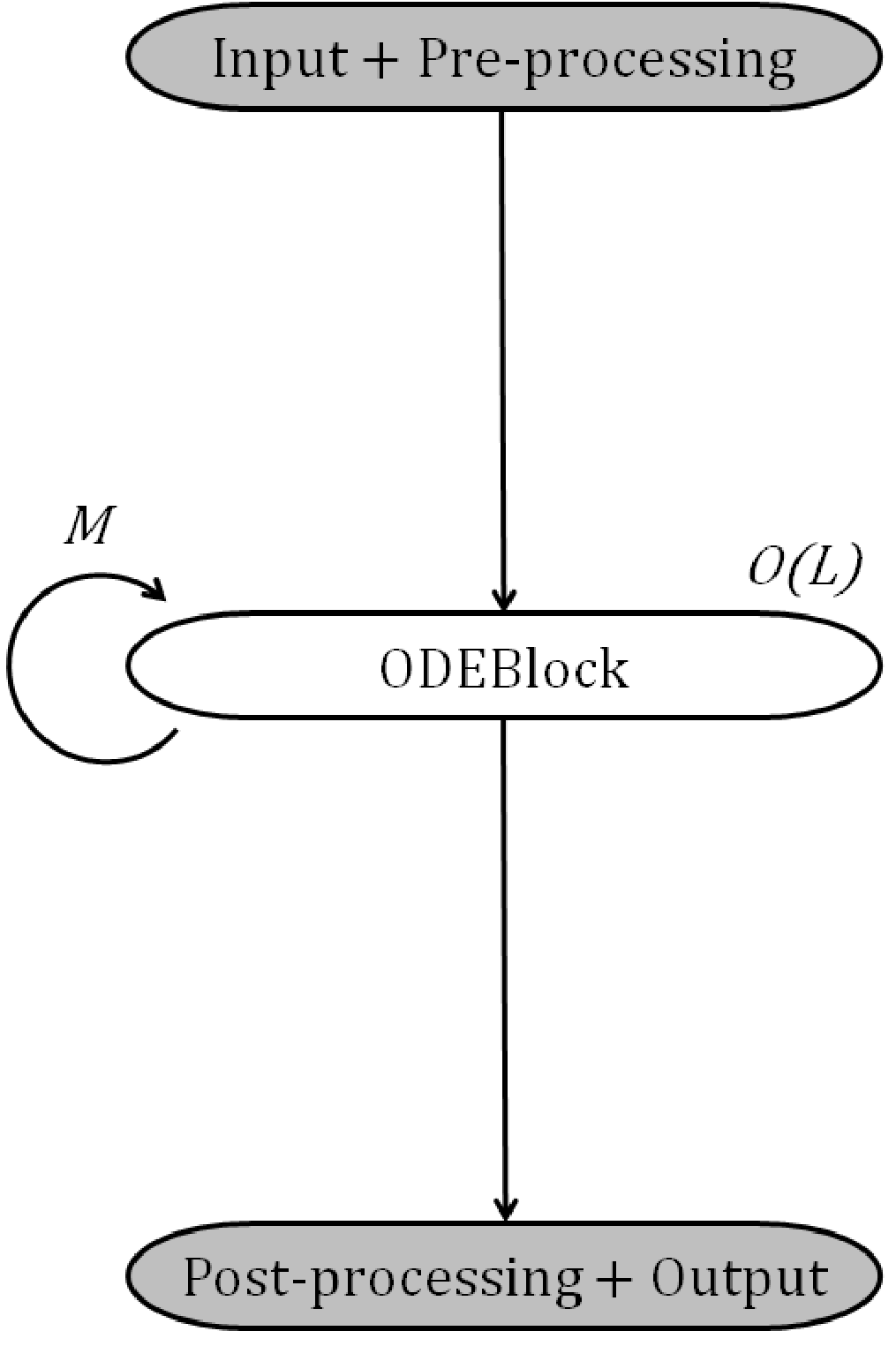}
			\caption{ODENet architecture}
			\label{fig:odenet}
		\end{minipage}
	\end{tabular}
\end{figure*}

ODENet \cite{ODENet} that employs an ordinary differential equation
(ODE) solver in DNNs was proposed to reduce weight
parameters of the network.
Stacking structure of layers in ResNet can be represented with an ODE
solver, such as Euler method.
ODENet thus uses an ODE solver in prediction and training processes so
that $M$ layers in ResNet are replaced with $M$ repeated executions of
a single layer, as shown in Figures \ref{fig:resnet} and \ref{fig:odenet}.
In this case, ODENet can significantly reduce the number of parameters
compared to the original ResNet while keeping the equivalent prediction
and training processes.

Field-Programmable Gate Array (FPGA) is an energy-efficient solution,
and it has been widely used in edge devices for machine learning
applications.
In this paper, we thus propose an FPGA-based acceleration of ODENet.
A core component of ODENet, called ODEBlock, that consists of
convolution layers, batch normalization \cite{batch_norm}, and
activation function is implemented on a programmable logic of low-cost
FPGA board, such as PYNQ-Z2 board.
Our contribution is that, as ODENet variants, reduced ODENets (rODENets) each
of which heavily uses a part of ODEBlocks and reduces/eliminates some layers
differently are proposed and analyzed for low-cost FPGA implementation.
They are evaluated in terms of parameter size, accuracy, execution
time, and resource utilization on the FPGA.

Many studies on FPGA-based DNN accelerators have been reported. 
In \cite{guo2018}, such accelerators and their techniques, such as
binarization and quantization, are surveyed. 
When a quantization using 2-bit weight parameters is applied to
ResNet-18, 10.3\% accuracy loss is reported.
In \cite{faraone2020}, circuit techniques that minimize information
loss from quantization are proposed and applied to ResNet-18 and 50.
In \cite{ma2017}, ResNet-50 and 152 are implemented on Intel Arria-10
FPGA using a 16-bit format and external memories.
Microsoft Brainwave platform supports ResNet-50 \cite{duarte2019}. 
Please note that this work focuses on low-cost FPGA platforms, such as
PYNQ-Z2.
Also our approach is orthogonal to quantization techniques and can 
be combined with for further reducing the parameter sizes.

The rest of this paper is organized as follows.
Section \ref{sec:prelim} provides a brief review of basic technologies
about ODENet.
Section \ref{sec:imple} implements a building block of ODENet on the
FPGA, and Section \ref{sec:eval} shows the evaluation results.
Section \ref{sec:conc} concludes this paper.

\section{Preliminaries}\label{sec:prelim}

\subsection{ResNet}\label{ssec:resnet}
In neural networks that serially stack many layers, training process
may be prevented when gradients become vanishingly small in one of the
layers (i.e., vanishing gradient problem).
Also, there may be a possibility that the training becomes unstable when
gradient descent is diverging (i.e., exploding gradient problem).
ResNet \cite{resnet} was proposed to address these issues and improve
the prediction accuracy by introducing shortcut connections that
enable to stack many layers.
Figure \ref{fig:resnet} illustrates ResNet architecture.
As shown in the figure, ResNet consists of a lot of building blocks.
Each building block receives input data $\bm{z}_t$ and executes
$3\times 3$ convolution, batch normalization \cite{batch_norm},
ReLU \cite{relu} as an activation function, $3\times 3$ convolution,
and batch normalization.
For example, ResNet architecture, which will be shown in Table
\ref{tb:network}, can be used in image classification tasks, such as
CIFAR-10 and CIFAR-100 datasets.
In this paper, building blocks executing the same computations are
grouped as layer$x$, where $x=1,2,3$.
ResNet size is denoted as ResNet-N, where N is the total number of
convolution and fully-connected steps in the building blocks including 
the pre- and post-processing layers.

Let $\bm{z}\in\mathbb{R}^{Z}$ and $\bm{y}\in\mathbb{R}^{Z}$ be an
input and an output of ResNet, respectively.
A network parameter $\theta$ is interpreted as a mapping function
$\mathcal{H}: \bm{z} \rightarrow \bm{y}$.
Assuming a normal forward propagation, an output of a building block
is represented as a function $f(\bm{z}, \theta)$.
When an input of a building block is additionally added to an output
of the building block with a shortcut connection, the function is
changed to $f(\bm{z}, \theta) + \bm{z}$.
Even with a shortcut connection, an output of the building block
itself is still $\mathcal{H}(\bm{z})$.
Thus, its residual should be trained in the training process so that
$f(\bm{z}, \theta) = \mathcal{H}(\bm{z}) - \bm{z}$.
In this case, a gradient at least contains 1; thus, vanishing gradient
problem can be mitigated.


Assuming ResNet consists of multiple building blocks, an input to the
$(t+1)$-th building block is represented as follows.
\begin{equation}\label{eq:resblock}
	\bm{z}_{t+1} = \bm{z}_t + f(\bm{z}_t, \theta_t),
\end{equation}
where $\bm{z}_t$ and $\theta_t$ denote the input and parameter of the
$t$-th building block, respectively.


\subsection{Ordinary Differential Equation}\label{ssec:ode}
ODE is an equation containing functions of one variable and their
derivatives.
For example, a first-order differential equation is represented as
follows.
\begin{equation}\label{eq:ode-1st}
	\frac{dz}{dt} = f(z(t), t, \theta),
\end{equation}
where $f$ and $\theta$ represent dynamics and the other parameters,
respectively.
Assuming $f$ is known and $z(t_0)$ is given, a problem to find
$z(t_1)$ that satisfies the above equation is known as an initial
value problem.
It is formulated as follows.
\begin{align}
	z(t_1) & = z(t_0) + \int_{t_0}^{t_1}{f(z(t), t, \theta)dt} \label{eq:ode-initial-con} \\
	       & = \textrm{ODESolve}(\bm{z}(t_0), t_0, t_1, f) \label{eq:ode-initial-dis}
\end{align}
In the right side of Equation \ref{eq:ode-initial-con}, the second
term contains an integral of a given function.
It cannot be solved analytically for arbitrary functions, so a
numerical approximation is typically employed to solve Equation
\ref{eq:ode-initial-con}.
To solve the equation, ODESolve function is defined as shown in
Equation \ref{eq:ode-initial-dis}.
In ODESolve function, integration range $[t_0, t_1]$ is divided into
partitions with step size $h$.
For $t_0 < \cdots t_i < \cdots t_1$, it computes corresponding $z_i$
using a recurrence formula.
As a method to compute $\bm{z}(t_1)$ in Equation \ref{eq:ode-initial-dis},
well-known ODE solvers, such as Euler method, second-order Runge-Kutta
method, and fourth-order Runge-Kutta method, can be used \cite{Press07}.
They can approximately solve Equation \ref{eq:ode-initial-con} in
the first-order, second-order, and fourth-order accuracy, respectively.
Below is Euler method.
\begin{equation}\label{eq:euler}
	z(t_{i+1}) = z(t_i) + hf(z(t_i), t_i, \theta)
\end{equation}



\subsection{ODENet}\label{ssec:ODENet}
An output of building blocks in ResNet can be computed with a
recurrence formula, as shown in Equation \ref{eq:resblock}.
Please note that Equation \ref{eq:resblock} is similar to Equation
\ref{eq:euler} except that the former basically assumes vector values
while the latter assumes scalar values.
Thus, one building block is interpreted as one step in Euler method.
As mentioned in Section \ref{ssec:ode}, since Euler method is a
first-order approximation of Equation \ref{eq:ode-initial-con},
an output of ResNet building block can be interpreted as well.
Since Equation \ref{eq:ode-initial-con} can be solved by Equation
\ref{eq:ode-initial-dis}, the output of ResNet can be solved by the
same equation.
Here, a building block of ResNet is replaced with an ODEBlock using
ODESolve function.
Neural network architecture consisting of such ODEBlocks is called
ODENet.
Figure \ref{fig:odenet} shows an ODENet architecture.
ODENet that repeats the same ODEBlock $M$ times is interpreted as
ResNet that implements $M$ building blocks.

Prediction tasks of ODENet are executed based on Equation
\ref{eq:ode-initial-dis}.
In training process, it is required that gradients are back-propagated
along neural network layers via ODESolve function.
To compute the gradients, ODENet uses an adjoint method \cite{adjoint}
in the training process.
Here, loss function $L$ of ODENet is represented as follows.
\begin{equation}\label{eq:loss_function}
	L(\bm{z}(t_1)) = L(\mathrm{ODESolve}(\bm{z}(t_0), t_0, t_1, f))
\end{equation}
Let an adjoint vector $\bm{a}$ be
$\bm{a} = \frac{\partial L }{\partial \bm{z}(t)}$.
The following equation is satisfied with respect to $\bm{a}$.
\begin{equation}\label{eq:da_dt}
	\frac{d\bm{a}(t)}{dt} = -\bm{a}^{\top}\frac{\partial f(\bm{z}(t), t, \theta)}{\partial \bm{z}(t)}
\end{equation}
Based on Equations \ref{eq:da_dt} and \ref{eq:ode-initial-dis}, a
gradient of parameter $\theta$ derived by a loss function is computed
as follows.
\begin{equation}\label{eq:grad}
	\begin{aligned}
		\frac{dL}{d\theta} & = -\int_{t_1}^{t_0}{\bm{a}(t)^{\top}\frac{\partial f(\bm{z}(t), t, \theta)}{\partial \theta}dt}                             \\
		                   & = \mathrm{ODESolve}\left(\bm{0}, t_1, t_0, -\bm{a}(t)^{\top}\frac{\partial f(\bm{z}(t), t, \theta)}{\partial \theta}\right) \\
	\end{aligned}
\end{equation}
$\bm{z}(t)$ and $\bm{a}(t)$ can be computed with ODESolve function.
A training function can be summarized as follows by using Equation
\ref{eq:grad} and these values \cite{ODENet}.
\begin{equation}\label{eq:augumented_dynamics}
	\begin{aligned}
		\bm{z}(t_0)        & = \mathrm{ODESolve}(\bm{z}(t_1), t_1, t_0, f(\bm{z}(t), t, \theta))                                                               \\
		\bm{a}(t_0)        & =\mathrm{ODESolve}\left(\bm{a}(t_1), t_1, t_0, -\bm{a}(t)^{\top}\frac{\partial f(\bm{z}(t), t, \theta}{\partial \bm{z}(t)}\right) \\
		\frac{dL}{d\theta} & = \mathrm{ODESolve}\left(\bm{0}, t_1, t_0, -\bm{a}(t)^{\top}\frac{\partial f(\bm{z}(t), t, \theta)}{\partial \theta}\right)
	\end{aligned}
\end{equation}
Please note that vector size of $\bm{0}$ in Equation
\ref{eq:augumented_dynamics} is same as that of $\theta$.
In the original Equation \ref{eq:grad}, it is necessary to compute
$a(t)$ and $z(t)$ for each $t$.
On the other hand, in the case of Equation \ref{eq:augumented_dynamics},
$z(t)$ is computed first using ODESolve function; then $a(t)$ is
computed based on $z(t)$, and the gradient is computed based on
$z(t)$ and $a(t)$.
Thus, the gradient is computed sequentially without keeping $a(t)$ and
$z(t)$ for each $t$, so memory usage can be reduced as well.
Based on the above-mentioned properties of ODENet, the benefit against
the original ResNet is that the number of parameters can be reduced
by ODENet.
In the prediction process, ResNet is represented with $M$ different
building blocks, while ODENet repeatedly uses a single ODEBlock $M$ times.
When the number of parameters for one building block is $O(L)$,
those of ResNet and ODENet are $O(LM)$ and $O(L)$, respectively.
As we will see in Table \ref{tb:resnet}, the number of parameters for
pre- and post-processing layers (e.g., conv1 and fc) is not
significant, it is expected that the number of parameters of ResNet is
reduced to approximately $\frac{1}{M}$.
In other words, different parameters
are used for each $t$ in ResNet, as shown in Equation \ref{eq:resblock}.
In ODENet, on the other hand, as shown in Equation \ref{eq:ode-1st},
$\theta$ is independent of $t$; thus, it can be trained while the
parameters are fixed irrespective of $t$.
Please note that different ODE solvers can be used in prediction and
training processes.
For example, a fourth-order Runge-Kutta method is used for training
with high accuracy, while Euler method is used for prediction tasks
for low latency and simplicity.
We can strike a balance between accuracy and performance by selecting
a proper solver.

\section{FPGA Implementation}\label{sec:imple}
\subsection{ODEBlock}\label{ssec:imple}

\begin{figure}[t!]
                \centering
                \includegraphics[height=57mm]{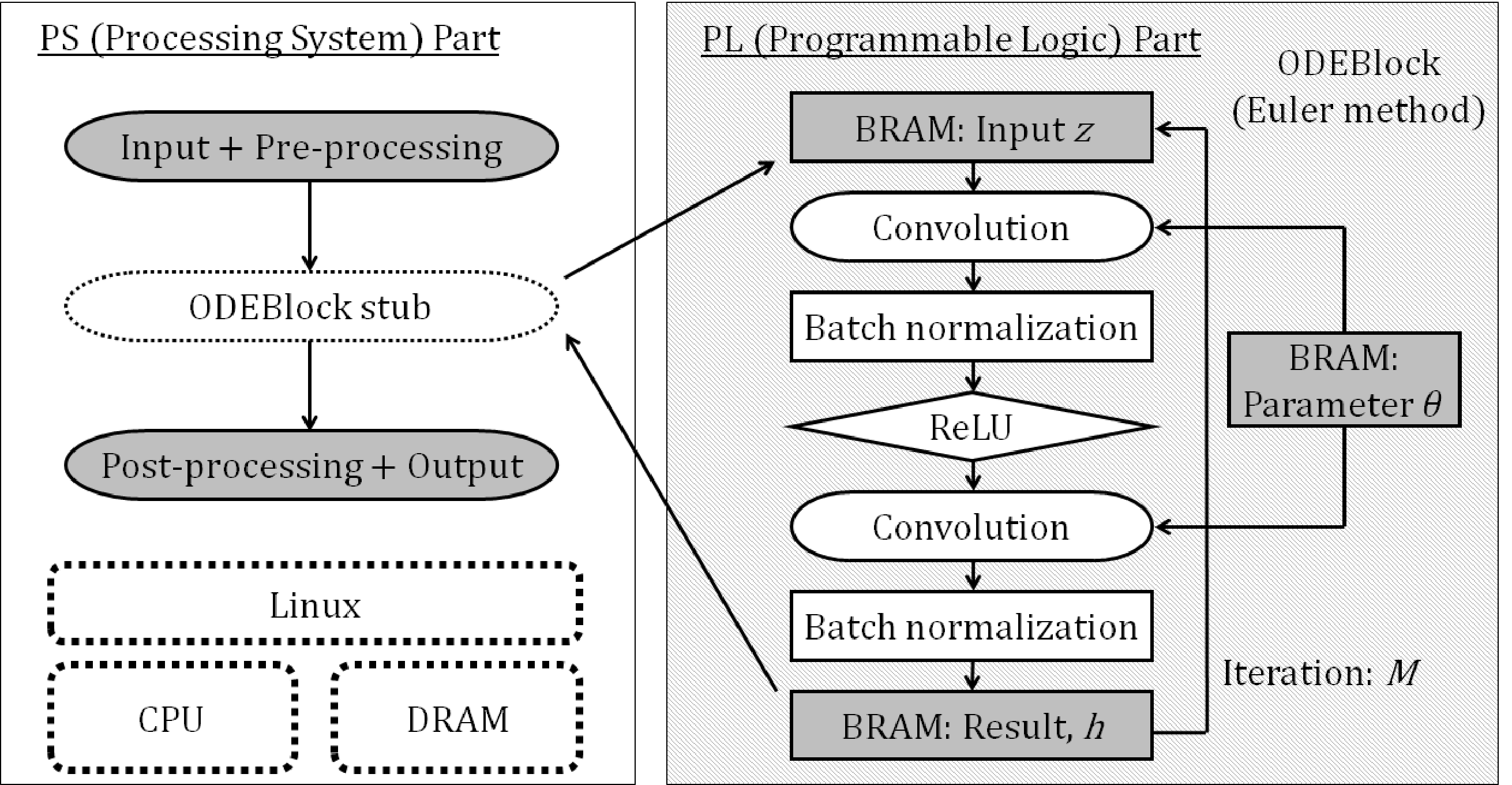}
                \caption{ODEBlock design on FPGA}
                \label{fig:design}
\end{figure}

\begin{figure}[htb]
\begin{minipage}{0.48\textwidth}
        \begin{center}
        \includegraphics[height=40mm]{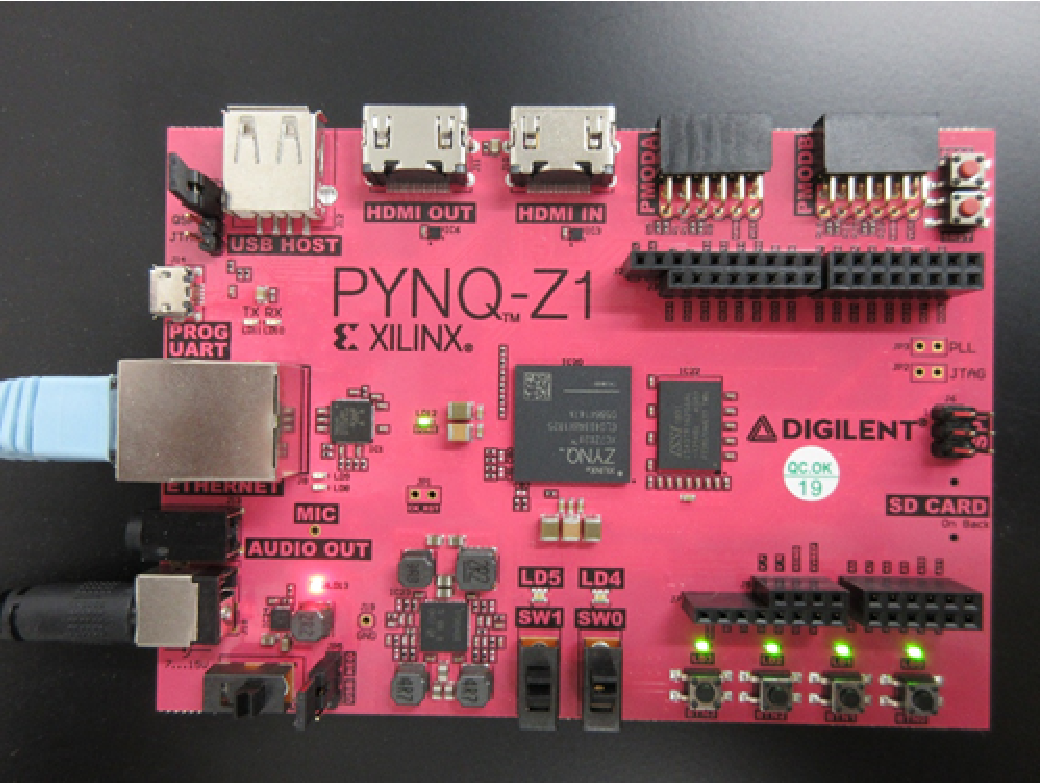}
        \caption{Overview of PYNQ-Z2 board}
        \label{fig:pynq}
        \end{center}
\end{minipage}
\begin{minipage}{0.5\textwidth}
        \begin{center}
        \makeatletter
        \def\@captype{table}
        \makeatother
        \caption{Specification of PYNQ-Z2 board}\vspace{3mm}
        \label{tb:fpga_env}
        \begin{tabular}{l|c} \hline
        OS   & PYNQ Linux (Ubuntu 18.04)          \\
        CPU  & ARM Cortex-A9 @ 650MHz $\times$ 2 \\
        DRAM & 512MB (DDR3)                      \\
        FPGA & Xilinx Zynq XC7Z020-1CLG400C      \\ \hline
        \end{tabular}
        \end{center}
\end{minipage}
\end{figure}

In this paper, as a target platform, we employ SoC type FPGA devices
that integrate programmable logic (PL) part and processor (PS) part,
as shown in Figure \ref{fig:design}.
PS part consists of CPU and DRAM, while PL part has programmable logic.
We use TUL PYNQ-Z2 board \cite{pynq} in this paper.
Figure \ref{fig:pynq} shows the FPGA board, and 
Table \ref{tb:fpga_env} shows the specification.
As shown in Figure \ref{fig:design}, a part of the ODEBlock is
implemented on PL part as a dedicated circuit, while the others are
executed on PS part as software.

\begin{table}[t!]
    \centering
    \caption{Network structure of ODENet}
    \label{tb:resnet}
    \begin{tabular}{c|c|c|c|c}\hline
        Layer                  & Output size                  & Detail                                & Parameter size [kB] & \# of executions per block   \\ \hline
        conv1                  & $32\times32$, 16ch           & $3\times3$, stride 1                  & 1.86      & 1               \\ \hline
        layer1                 & $32\times32$, 16ch           & $\begin{bmatrix}3\times3 \\ 3\times3\end{bmatrix}$, stride 1 & 19.84     & $\frac{N-2}{6}$ \\ \hline
        layer2\_1              & $16\times16$, 32ch           & $\begin{bmatrix}3\times3 \\ 3\times3\end{bmatrix}$, stride 2 & 55.81     & 1               \\ \hline
        layer2\_2              & $16\times16$, 32ch           & $\begin{bmatrix}3\times3 \\ 3\times3\end{bmatrix}$, stride 1 & 76.54     & $\frac{N-8}{6}$ \\ \hline
        layer3\_1              & $8\times8$, 64ch             & $\begin{bmatrix}3\times3 \\ 3\times3\end{bmatrix}$, stride 2 & 222.21    & 1               \\ \hline
        layer3\_2              & $8\times8$, 64ch             & $\begin{bmatrix}3\times3 \\ 3\times3\end{bmatrix}$, stride 1 & 300.54    & $\frac{N-8}{6}$ \\ \hline
        fc                     & $1\times100$                 & \begin{tabular}{@{}c@{}}Average pooling,\\100d fc, softmax\end{tabular}             & 26.00     & 1               \\ \hline
    \end{tabular}
\end{table}

Table \ref{tb:resnet} shows network structure of ODENet with a given N.
It consists of several building blocks or ``layers'' as shown in the
table: conv1, input1, input2\_1, input2\_2, input3\_1, input3\_2, and
fc.
In the case of ODENet, only a single block instance is implemented for
each layer, and the same instance is continuously executed instead.
For example, the numbers of executions per block for input1, input2\_2, and
input3\_2 are $\frac{N-2}{6}$, $\frac{N-8}{6}$, and $\frac{N-8}{6}$,
respectively; the others are executed only once.
In this paper, we thus implement input1, input2\_2, and input3\_2
individually on a resource-limited FPGA board.
That is, each of these layers is implemented on PL part of PYNQ-Z2,
while the other parts are executed on PS part as software.
Euler method is used as an ODE solver.

Each layer consists of five steps:
1) convolution,
2) batch normalization,
3) activation function (ReLU),
4) convolution, and
5) batch normalization.
The convolution step differs in each layer.
That is, the input/output channel numbers for layer1, layer2\_2, and
layer3\_2 are 16, 32, and 64.
The input/output feature map sizes for layer1, layer2\_2, and
layer3\_2 are 8$\times$8, 16$\times$16, and 32$\times$32.
Their kernel size is 3$\times$3 and stride width is 1.
The above mentioned five steps are implemented in Verilog HDL.
32-bit Q20 fixed-point number format is used.
Multiply-add units are used in the convolution and ReLU steps, and
multiply-add units, division unit, and square root unit are used in
the batch normalization steps for computing mean, variance, and standard
deviation.
Weight parameters $\theta$ of the two convolution steps are stored in
Block RAM (BRAM) of the FPGA.
Input and output feature maps for all the channels are also stored in
the BRAM.

Most of computation time is consumed in the convolution steps
\footnote{
    The two convolution steps consume about 99\% of execution cycles
    of layer3\_2 when only a single multiply-add unit is used in our
    implementation.
}.
Our convolution and ReLU step implementations are scalable; that is,
we can increase the number of multiply-add units from 1 to 64
depending on available resources but it is also restricted by the
number of output channels.
Their execution cycles (except for the batch normalization) decrease
in inverse proportion to the number of multiply-add units.
We implemented layer1, layer2\_2, and layer3\_2 each using 1, 4, 8,
16, and 32 multiply-add units.
They are referred to as conv\_x1, conv\_x4, conv\_x8, conv\_x16, and
conv\_x32, respectively.
For example, their execution cycles of layer3\_2 are 23.78M, 6.07M,
3.12M, 1.64M, and 0.90M cycles, respectively.
In these implementations, since only conv\_x32 could not satisfy a
timing constraint of our target FPGA board (i.e., 100MHz), we mainly
use conv\_x16 in this paper.

\begin{table}[th]
    \centering
    \caption{Resource utilizations of layer1, layer2\_2, and layer3\_2 on Zynq XC7Z020}
    \label{tb:resource}
    \begin{tabular}{c|c|rr|rr|rr|rr}\hline
        Layer                      & Parallelism & \multicolumn{2}{|c|}{BRAM} & \multicolumn{2}{|c|}{DSP} & \multicolumn{2}{|c|}{LUT} & \multicolumn{2}{|c}{FF}                                       \\ \hline
        \multirow{4}{*}{layer1}    & conv\_1     & 56                         & (40.00\%)                 & 8                         & (3.63\%)                & 1486  & (2.79\%)  & 835  & (0.78\%) \\
                                   & conv\_4     & 56                         & (40.00\%)                 & 20                        & (9.09\%)                & 2992  & (5.62\%)  & 1358 & (1.28\%) \\
                                   & conv\_8     & 56                         & (40.00\%)                 & 36                        & (16.36\%)               & 4740  & (8.91\%)  & 2058 & (1.93\%) \\
                                   & conv\_16    & 64                         & (45.71\%)                 & 68                        & (30.91\%)               & 8994  & (16.91\%) & 4145 & (3.90\%) \\ \hline
        \multirow{4}{*}{layer2\_2} & conv\_1     & 56                         & (40.00\%)                 & 8                         & (3.63\%)                & 1482  & (2.79\%)  & 833  & (0.78\%) \\
                                   & conv\_4     & 56                         & (40.00\%)                 & 20                        & (9.09\%)                & 2946  & (5.53\%)  & 1346 & (1.27\%) \\
                                   & conv\_8     & 56                         & (40.00\%)                 & 36                        & (16.36\%)               & 4737  & (8.90\%)  & 2032 & (1.91\%) \\
                                   & conv\_16    & 56                         & (40.00\%)                 & 68                        & (30.91\%)               & 8844  & (16.62\%) & 4873 & (4.58\%) \\ \hline
        \multirow{4}{*}{layer3\_2} & conv\_1     & 140                        & (100.00\%)                & 8                         & (3.63\%)                & 1692  & (3.18\%)  & 927  & (0.87\%) \\
                                   & conv\_4     & 140                        & (100.00\%)                & 20                        & (9.09\%)                & 3048  & (5.73\%)  & 1411 & (1.33\%) \\
                                   & conv\_8     & 140                        & (100.00\%)                & 36                        & (16.36\%)               & 4907  & (9.22\%)  & 2059 & (1.94\%) \\
                                   & conv\_16    & 140                        & (100.00\%)                & 68                        & (30.91\%)               & 12720 & (23.91\%) & 6378 & (5.99\%) \\ \hline
    \end{tabular}
\end{table}

\subsection{Resource Utilization}\label{ssec:resource}
Table \ref{tb:resource} shows resource utilizations of layer1,
layer2\_2, and layer3\_2 implemented on PL part of the FPGA for ODENet
and its variants in this paper.
Here, we show the result when $n$ multiply-add units are used for
the convolution and ReLU steps.
They are denoted as conv\_x$n$ implementations.
As shown in the table, if we implement layer3\_2 on PL part of the
FPGA, BRAM utilization becomes 100\%.
In this case, the utilizations of DSP, LUT, and FF still have room and
can implement some other application logic, but we cannot implement
more weight parameters or larger feature maps without relying on
external DRAMs.
On the other hand, BRAM utilizations of layer1 and layer2\_2 are not
as high as layer3\_2, and the other resources also have enough room,
so we can implement both the layers on PL part of the FPGA.
In the next section, we can thus consider four cases: 
1) only layer1 is implemented on PL part,
2) only layer2\_2 is implemented on PL part,
3) layer1 and layer2\_2 are implemented on PL part, and
4) only layer3\_2 is implemented on PL part of the FPGA.

\section{Evaluations}\label{sec:eval}
CIFAR-100 is used as a dataset in this paper.
ODENet on the FPGA is evaluated in terms of the number of parameters,
accuracy, and execution time when a part of convolution layers is
executed by PL part.

\subsection{Network Configuration}\label{ssec:network}
Here, we introduce reduced ODENet (rODENet) variants for low-cost
FPGA implementation.
As shown in Table \ref{tb:network}, seven network architectures
including our rODENet variants listed
below are used in this evaluation.
Please note that the number of stacked blocks means the number of
block instances implemented, while the number of executions per block
means the number of iterations on the same block instance.

\begin{itemize}
	\item {\bf ResNet}-N: Baseline ResNet
	      	      	      	      	      	      	      
	\item {\bf ODENet}-N:
	      layer1, layer2\_2, and layer3\_2 in {\bf ResNet}-N are replaced with
	      corresponding ODEBlocks.
	      	      	      	      	      	      	      
	\item {\bf rODENet-1}-N:
	      layer2\_2 and layer3\_2 are removed.
	      layer1 is replaced with ODEBlock, and the number of executions on
	      layer1 is increased instead so that the total execution count of
	      building blocks is same as {\bf ResNet}-N.
	\item {\bf rODENet-2}-N:
	      The number of executions on layer1 is reduced to 1 and layer3\_2 is removed.
	      layer2\_2 is replaced with ODEBlock, and the number of executions on
	      layer2\_2 is increased instead.
	\item {\bf rODENet-1+2}-N:
	      layer3\_2 is removed.
	      layer1 and layer2\_2 are replaced with ODEBlocks, and the numbers of
	      executions on layer1 and layer2\_2 are increased instead.
	\item {\bf rODENet-3}-N:
	      The number of executions on layer1 is reduced to 1 and layer2\_2 is removed.
	      layer3\_2 is replaced with ODEBlock, and the number of executions on
	      layer3\_2 is increased instead.
	      	      	      	      	      	      	      
	\item {\bf Hybrid-3}-N:
	      Only layer3\_2 in {\bf ResNet}-N is replaced with ODEBlock.
	      The other layers are the same as those in ResNet.
\end{itemize}

We can expect that a computation of {\bf ODENet}-N is compatible with
that in {\bf ResNet}-N.
On the other hand, our {\bf rODENet-1}-N, {\bf rODENet-2}-N,
{\bf rODENet-1+2}-N, and {\bf rODENet-3}-N execute the same number of
building blocks as {\bf ResNet}-N, but they heavily use layer1,
layer2\_2, layer1 and layer2\_2, and layer3\_2, respectively.
Our intention is that these heavily-used layers are offloaded to
PL part as shown in Figure \ref{fig:design}.
In addition, {\bf Hybrid-3}-N, which is a middle of {\bf ResNet}-N
and {\bf ODENet}-N, is evaluated as a high-accuracy variant.

Euler method is used as an ODE solver, because it is simple and
requires only a small temporary memory at prediction time.
In Table \ref{tb:network}, conv1 is the pre-processing step that
executes 3$\times$3 convolution, batch normalization, and ReLU as an
activation function.
Then, various building blocks (e.g., layer1 to layer3\_2 in Table
\ref{tb:network}) are executed as shown in Figure \ref{fig:resnet}.
Finally, fc is the post-processing step that executes global average
pooling, fully-connected layer to all the output classes, and Softmax
as an activation function.
Stride width is set to 1 in most of building blocks except for
layer2\_1 and layer3\_1, in which stride width is set to 2 in order to
reduce the output feature map size.

\begin{table}[t!]
  \centering
  \caption{Network structure of ResNet, ODENet, and rODENet variants}
  \label{tb:network}
  \begin{tabular}{c|c|c|c|c|c|c|c}\hline
    \multirow{2}{*}{Layer} & \multicolumn{7}{c}{\# of stacked blocks / \# of executions per block}                                                                                                                                     \\ \cline{2-8}
                           & {\bf ResNet}                                                          & {\bf ODENet}        & {\bf rODENet-1}     & {\bf rODENet-2}     & {\bf rODENet-1+2}   & {\bf rODENet-3}     & {\bf Hybrid-3}      \\ \hline
    conv1                  & 1 / 1                                                                 & 1 / 1               & 1 / 1               & 1 / 1               & 1 / 1               & 1 / 1               & 1 / 1               \\ \hline
    layer1                 & $\frac{N-2}{6}$ / 1                                                   & 1 / $\frac{N-2}{6}$ & 1 / $\frac{N-6}{2}$ & 1 / 1               & 1 / $\frac{N-4}{4}$ & 1 / 1               & $\frac{N-2}{6}$ / 1 \\ \hline
    layer2\_1              & 1 / 1                                                                 & 1 / 1               & 1 / 1               & 1 / 1               & 1 / 1               & 1 / 1               & 1 / 1               \\ \hline
    layer2\_2              & $\frac{N-8}{6}$ / 1                                                   & 1 / $\frac{N-8}{6}$ & 0 / 0               & 1 / $\frac{N-8}{2}$ & 1 / $\frac{N-8}{4}$ & 0 / 0               & $\frac{N-8}{6}$ / 1 \\ \hline
    layer3\_1              & 1 / 1                                                                 & 1 / 1               & 1 / 1               & 1 / 1               & 1 / 1               & 1 / 1               & 1 / 1               \\ \hline
    layer3\_2              & $\frac{N-8}{6}$ / 1                                                   & 1 / $\frac{N-8}{6}$ & 0 / 0               & 0 / 0               & 0 /0                & 1 / $\frac{N-8}{2}$ & 1 / $\frac{N-8}{6}$ \\ \hline
    fc                     & 1 / 1                                                                 & 1 / 1               & 1 / 1               & 1 / 1               & 1 / 1               & 1 / 1               & 1 / 1               \\ \hline
  \end{tabular}
  \centering
\end{table}

\subsection{Parameter Size}\label{ssec:param_size}
\begin{figure*}[t!]
	\centering
	\includegraphics[height=88.5mm]{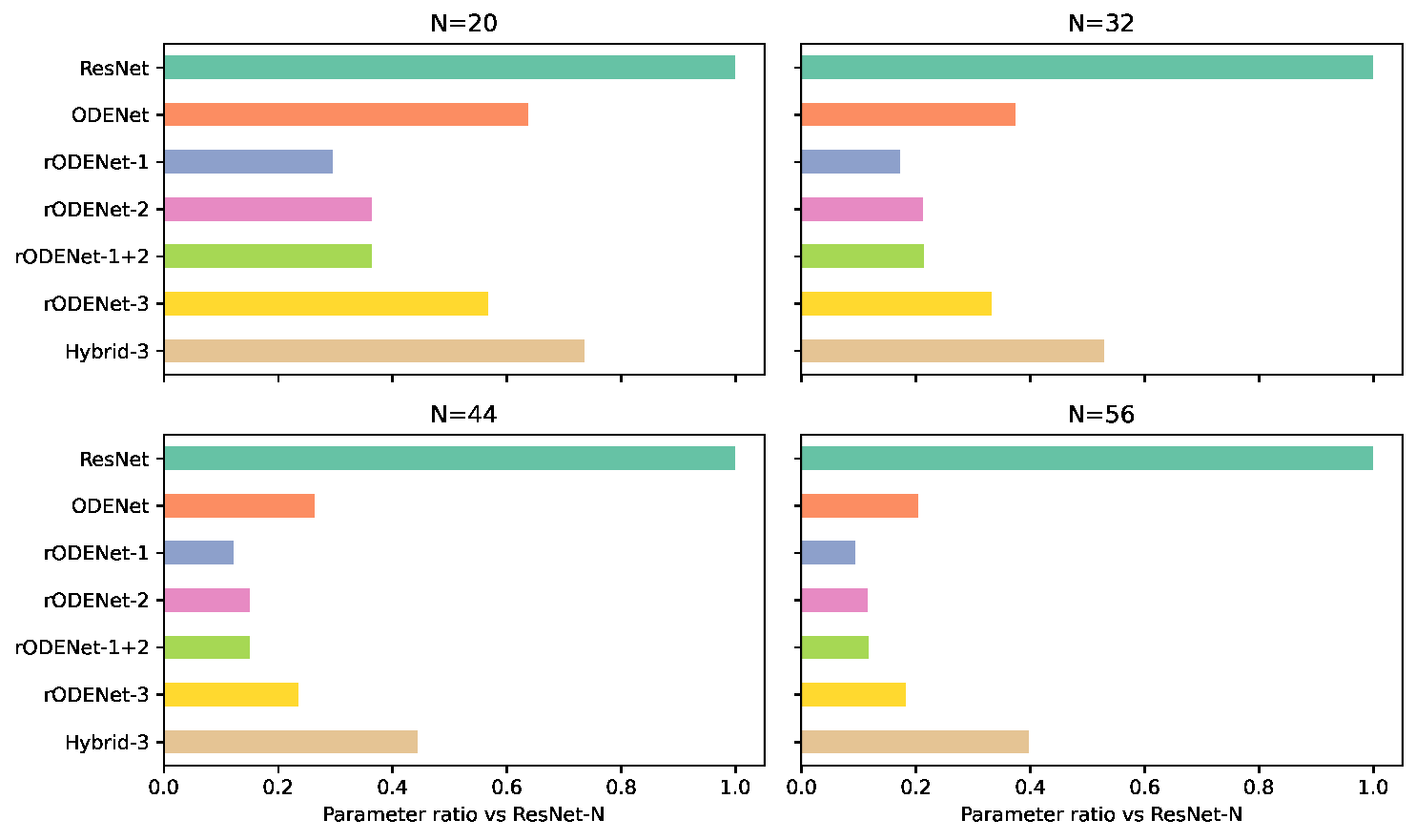}
	\caption{Parameter size of ResNet, ODENet, and rODENet variants}
	\label{fig:parameters}
\end{figure*}

Figure \ref{fig:parameters} shows the total parameter size for each
architecture listed in Section \ref{ssec:network}, assuming that each
parameter is implemented in a 32-bit format.

As shown in Table \ref{fig:parameters}, parameter size of
{\bf ResNet}-N is proportional to the number of stacked building
blocks (see Figure \ref{fig:resnet}).
Please note that parameter sizes of {\bf ODENet}-N and the
rODENet variants are independent of N, since the number of
stacked instances is independent of N (see Figure \ref{fig:odenet}).
In the rODENet variants, their parameter sizes depend on
layers actually implemented.
When N is 20 (the smallest case), parameter sizes of {\bf ODENet}-N
and {\bf rODENet-3} are 36.24\% and 43.29\% less than that of
{\bf ResNet}-20, respectively.
When N is 56 (the largest case), their parameter sizes are
79.54\% and 81.80\% less than that of {\bf ResNet}-56, respectively.
Although structure of {\bf Hybrid-3}-N is similar to that of
{\bf ResNet}-N except for layer3\_2, it can reduce the parameter size
by 26.43\% and 60.16\% compared to {\bf ResNet}-N when N is 20 and 56,
respectively.
Please note that the parameter size reduction by using ODEBlock is
independent of the other parameter reduction techniques, such as
quantization \cite{guo2018}, and can be incorporated with them to
further reduce the parameter size.

\subsection{Accuracy}\label{ssec:accuracy}
In this experiment, SGD \cite{sgd} is used as an optimization
function.
As L2 regularization, $1\times 10^{-4}$ is added to each layer.
For the training process, the number of epochs is 200.
The learning rate is started with 0.01, and it is reduced by
$\frac{1}{10}$ when the epoch becomes 100 and 150.

\begin{figure*}[t!]
	\centering
	\includegraphics[height=102mm]{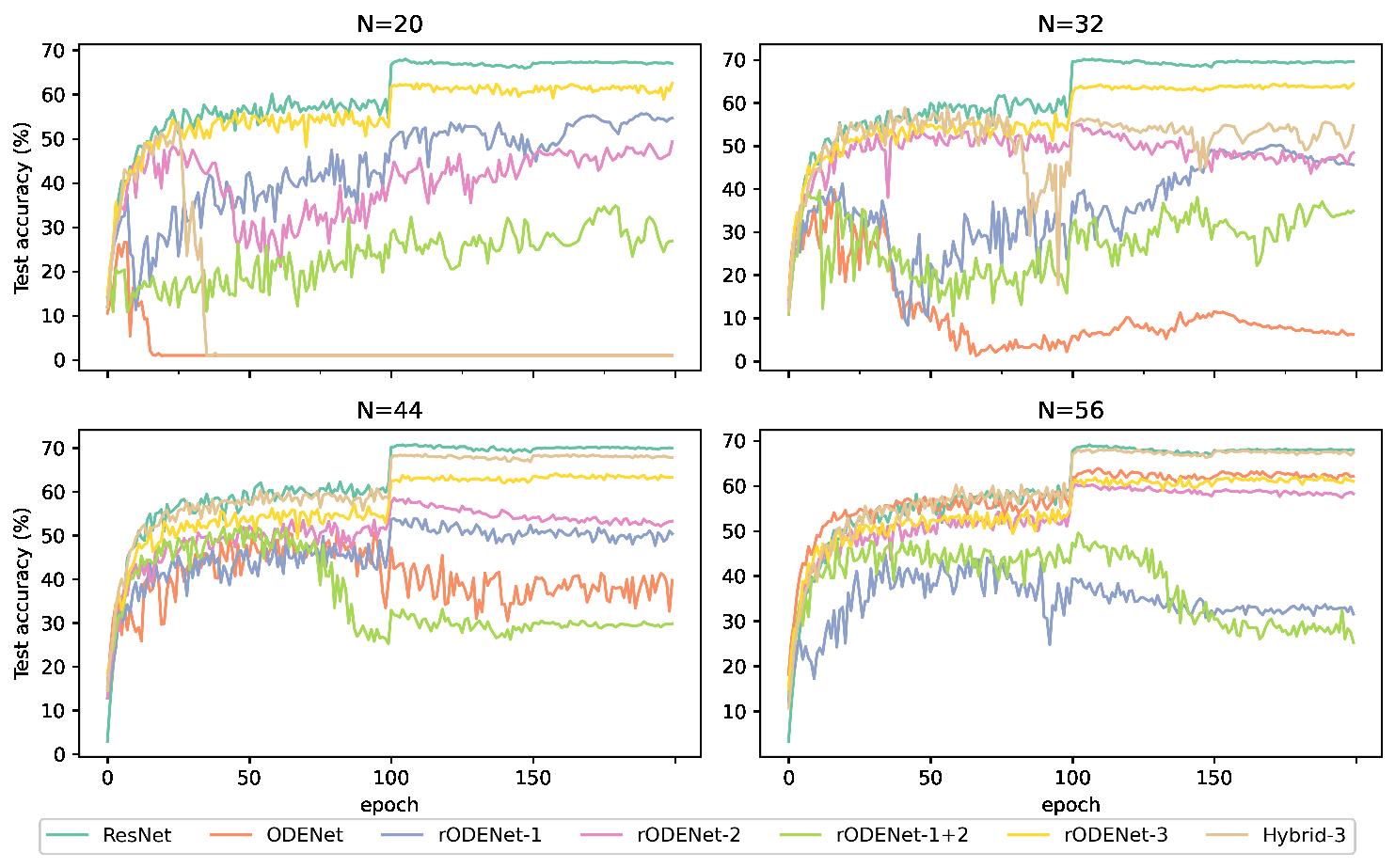}
	\caption{Accuracy of four network architectures when N=\{20,32,44,56\}}
	\label{fig:accuracy}
\end{figure*}

Figure \ref{fig:accuracy} shows the evaluation results of accuracy in
the seven network architectures listed in Table \ref{tb:network}.
As shown in the graphs, when N is 20, the training results are
unstable especially in {\bf ODENet}-N and {\bf Hybrid-3}-N.
When N is 56, on the other hand, accuracies of most
architectures are improved and become stable, except for
{\bf rODENet-1}-N and {\bf rODENet-1+2}-N.
Among the rODENet variants, {\bf rODENet-3} is stable and
shows relatively high accuracy when N is 20, 32, 44, and 56;
thus we mainly focus on {\bf rODENet-3}.

In {\bf Hybrid-3}-N, the accuracy is the almost same as
{\bf ResNet}-N when N is 44 and 56.
More specifically, accuracies of {\bf ResNet}-44 and {\bf Hybrid-3}-44
are 70.74\% and 68.58\%, and those of {\bf ResNet}-56 and
{\bf Hybrid-3}-56 are 69.09\% and 68.11\%; thus there
is up to 2.16\% accuracy loss.
The accuracy difference between {\bf ResNet}-44 and {\bf ResNet}-56 is
1.65\%, while that of {\bf Hybrid-3}-44 and {\bf Hybrid-3}-56 is
only 0.47\%; thus, they are robust against overfitting (i.e.,
degradation of generalization ability) due to larger N.

In {\bf rODENet-3}-N, the accuracy is the second highest next to
that of {\bf ResNet}-N when N is 20 and 32.
Accuracies of {\bf ResNet}-20 and {\bf rODENet-3}-20 are 68.02\% and
62.54\%, and the accuracy difference is 5.48\%.
Those of {\bf ResNet}-32 and {\bf rODENet-3}-32 are 70.16\% and
64.46\%, and the difference is 5.70\%.
These differences are large compared to those of {\bf Hybrid-3}-N
when N is 44 and 56.
Still, {\bf rODENet-3}-N is stable for all the sizes,
which means that this architecture has the highest robustness against
increasing N among the rODENet variants.
Because of its stability, we can use {\bf rODENet-3}-N even if
the optimal network architecture is not known yet.

In {\bf ODENet}-N, the accuracy is relatively high next to
those of {\bf ResNet}-N and {\bf Hybrid-3}-N when N is 56.
However, it is unstable when N is small.
The reason for this unstability is that the step size was relatively large and
thus it could not acquire the dynamics sufficiently.
In ODENet and its variants, it is interpreted that connections of
ResNet layers are continuous.
It is pointed out that there may be a possibility that ODENet cannot
compute the gradients accurately \cite{anode}.
This may be one reason for the unstability when N is small.

In summary, our {\bf rODENet-3} is stable and shows relatively high
accuracy for all the sizes in our experiment.
Considering the parameter size, it can strike a balance between the
parameter size reduction, stability, and accuracy.


\subsection{Execution Time}\label{ssec:exec_time}
The ODENet variants implemented on the FPGA mentioned in Section
\ref{sec:imple} are evaluated in terms of prediction time.
More specifically, the following architectures are compared.

\begin{itemize}
	\item {\bf ResNet}-N:
	      In ResNet-N, all the layers are executed on PS part as software.
	      	      	      	      	      	      	      
	\item {\bf rODENet-*}-N:
	      In the rODENet variants, only heavily-used layers are offloaded
	      to PL part as dedicated circuits. The other layers are software.
	      Among rODENet variants, we mainly focus on {\bf rODENet-3}-N
	      since it is advantageous in terms of the parameter size and accuracy.
	      In this case, layer3\_2 is PL part and the others are PS part.
	      	      	      	      	      	      	      
	\item {\bf ODENet-3}-N:
	      In ODENet-N, layer3\_2 is PL part and the others are PS part.
	      	      	      	      	      	      	      
	\item {\bf Hybrid-3}-N:
	      In Hybrid-3-N, layer3\_2 is PL part and the others are PS part.
\end{itemize}

An image size in CIFAR-100 dataset is (channel, height, width) = (3,
32, 32), and prediction time for each image is measured.
As an FPGA platform, TUL PYNQ-Z2 board 
that integrates PS and PL parts is used in this experiment.
As listed in Table \ref{tb:fpga_env}, in PS part, two ARM Cortex-A9
processors are running at 650MHz.
In PL part, the operating frequency of the dedicated circuits is 100MHz.
Vivado 2017.2 was used for the design synthesis and implementation of
the ODEBlocks (i.e., layer1, layer2\_2, and layer3\_2) implemented on
PL part.
We employ conv\_x16 implementation that uses 16 multiply-add units for
the convolution and ReLU steps.
PS and PL parts are typically connected via AXI bus and DMA transfer
is used for their communication though not fully implemented in our design.
We assume that data transfer latency between PS and PL parts is 1
cycle per float32.
This is an optimistic assumption, but we use this value for simplicity
because it varies depending on an underlying hardware platform.

Table \ref{tb:time} shows the execution times and speedup rates of the
seven architectures mentioned above when they are implemented on the
FPGA board.
In this table, ``Offload target'' means layer(s) implemented on PL
part of the FPGA.
For example, the offloaded target is layer3\_2 in {\bf rODENet-3}-N,
{\bf ODENet-3}-N, and {\bf Hybrid-3}-N.
As shown in the table, execution time of layer3\_2 takes up only
21.24\% to 29.64\% of total execution time of {\bf ODENet-3}-N and
{\bf Hybrid-3}-N.
On the other hand, layer3\_2 is heavily used intentionally in
{\bf rODENet-3}-N, and its execution time takes up 64.48\% to 87.87\%.
Thus, by offloading layer3\_2 to PL part, the total execution time of
{\bf rODENet-3}-N is 2.66 times faster than a pure software execution
when N is 56, which is the largest overall speedup by the FPGA.

\begin{table}[t!]
	\caption{Execution time of ResNet, ODENet, and rODENet variants
	(PS: Cortex-A9 @650MHz, PL: @100MHz)}
	\label{tb:time}
\hspace{-7mm}
	\begin{tabular}{c|c|c|c|c|c|c|c|c} \hline
		Model & N & \begin{tabular}{c} 
		Offload \\target
	\end{tabular} & \begin{tabular}{c}
	Total w/o \\PL [s]
	\end{tabular}          & \begin{tabular}{c}
	Target w/o \\PL [s]
	\end{tabular} & \begin{tabular}{c}
	Ratio of \\target [\%]\\
	\end{tabular} & \begin{tabular}{c}
	Target w/ \\PL [s]
	\end{tabular} & \begin{tabular}{c}
	Total w/ \\PL [s]
	\end{tabular} & \begin{tabular}{c}
	Overall \\speedup
	\end{tabular}        \\ \hline
	\multirow{4}{*}{{\bf ResNet}}      & 20                       & \multirow{4}{*}{--}                 & 0.54                       & --                         & --                         & --                         & --                         & --   \\
	& 32                       &                                     & 0.89                       & --                         & --                         & --                         & --                         & --   \\
	& 44                       &                                     & 1.24                       & --                         & --                         & --                         & --                         & --   \\
	& 56                       &                                     & 1.58                       & --                         & --                         & --                         & --                         & --   \\ \hline
							
	\multirow{4}{*}{{\bf rODENet-1}}   & 20                       & \multirow{4}{*}{layer1}             & 0.57                       & 0.44                       & 76.89                    & 0.15                       & 0.28                       & 1.99 \\
	& 32                       &                                     & 0.94                       & 0.81                       & 86.06                    & 0.29                       & 0.42                       & 2.26 \\
	& 44                       &                                     & 1.30                       & 1.17                       & 89.91                    & 0.42                       & 0.55                       & 2.37 \\
	& 56                       &                                     & 1.67                       & 1.54                       & 92.14                    & 0.55                       & 0.68                       & 2.45 \\ \hline
							
	\multirow{4}{*}{{\bf rODENet-2}}   & 20                       & \multirow{4}{*}{layer2\_2}          & 0.52                       & 0.33                       & 63.82                    & 0.11                       & 0.30                       & 1.75 \\
	& 32                       &                                     & 0.86                       & 0.67                       & 77.74                    & 0.22                       & 0.41                       & 2.08 \\
	& 44                       &                                     & 1.19                       & 1.00                       & 84.14                    & 0.33                       & 0.52                       & 2.28 \\
	& 56                       &                                     & 1.52                       & 1.33                       & 87.46                    & 0.44                       & 0.63                       & 2.40 \\ \hline
							
	\multirow{4}{*}{{\bf rODENet-1+2}} & 20                       &                                     & 0.55                       & 0.25 / 0.17                & 44.98 / 31.09          & 0.09 / 0.06                & 0.27                       & 1.99 \\
	& 32                       & layer1 /                            & 0.89                       & 0.42 / 0.33                & 47.54 / 37.71          & 0.15 / 0.11                & 0.39                       & 2.24 \\
	& 44                       & layer2\_2                           & 1.23                       & 0.60 / 0.50                & 48.63 / 40.75          & 0.22 / 0.17                & 0.52                       & 2.38 \\
	& 56                       &                                     & 1.60                       & 0.81 / 0.66                & 50.40 / 41.45          & 0.29 / 0.22                & 0.64                       & 2.52 \\ \hline
							
	\multirow{4}{*}{{\bf rODENet-3}}   & 20                       & \multirow{4}{*}{layer3\_2}          & 0.54                       & 0.35                       & 64.48                    & 0.10                       & 0.29                       & 1.85 \\
	& 32                       &                                     & 0.88                       & 0.69                       & 78.44                    & 0.20                       & 0.39                       & 2.26 \\
	& 44                       &                                     & 1.23                       & 1.04                       & 84.44                    & 0.30                       & 0.49                       & 2.50 \\
	& 56                       &                                     & 1.57                       & 1.38                       & 87.87                    & 0.40                       & 0.59                       & 2.66 \\ \hline
							
	\multirow{4}{*}{{\bf ODENet-3}}    & 20                       & \multirow{4}{*}{layer3\_2}          & 0.56                       & 0.12                       & 21.24                    & 0.03                       & 0.47                       & 1.18 \\
	& 32                       &                                     & 0.90                       & 0.23                       & 25.83                    & 0.07                       & 0.74                       & 1.23 \\
	& 44                       &                                     & 1.25                       & 0.34                       & 27.67                    & 0.10                       & 1.00                       & 1.24 \\
	& 56                       &                                     & 1.60                       & 0.46                       & 28.98                    & 0.13                       & 1.27                       & 1.26 \\ \hline
							
	\multirow{4}{*}{{\bf Hybrid-3}}    & 20                       & \multirow{4}{*}{layer3\_2}          & 0.53                       & 0.12                       & 22.38                    & 0.03                       & 0.44                       & 1.19 \\
	& 32                       &                                     & 0.88                       & 0.23                       & 26.65                    & 0.07                       & 0.71                       & 1.24 \\
	& 44                       &                                     & 1.23                       & 0.35                       & 28.11                    & 0.10                       & 0.99                       & 1.25 \\
	& 56                       &                                     & 1.56                       & 0.46                       & 29.64                    & 0.13                       & 1.23                       & 1.27 \\ \hline
	\end{tabular}
\end{table}

Regarding {\bf Hybrid-3}-N and {\bf ODENet-3}-N, the overall speedup
by the FPGA for {\bf Hybrid-3}-N is equal to or higher than that of {\bf ODENet-3}-N
in all the sizes.
This is because the ratio of layer3\_2 in {\bf Hybrid-3}-N is slightly
higher than that in {\bf ODENet-3}-N and the speedup rate of only
layer3\_2 by the FPGA is almost constant regardless of N.

In summary, the overall speedup rate by the FPGA is relatively high in
the rODENet variants, followed by {\bf Hybrid-3}-N and {\bf ODENet-3}-N.
Although all the rODENet variants show favorable speedup, only
{\bf rODENet-3}-N shows high and stable accuracy, as shown in Section
\ref{ssec:accuracy}.
Regarding the overall speedup compared to the original ResNet,
{\bf rODENet-3}-56 is 2.67 times faster than a pure software
execution of {\bf ResNet}-56.
Although the overall speedup by the FPGA is smallest in
{\bf Hybrid-3}-20, it is still 1.22 times faster than a software
execution of {\bf ResNet}-20.
Please note that, as mentioned in Section \ref{ssec:accuracy}, the
accuracy of {\bf rODENet-3}-N is quite high and stable when N is 20
and 32.
When N is 44 and 56, its accuracy is less than {\bf Hybrid-3}-N, but
it is still comparable to {\bf ODENet-3}-N.
Thus, our proposed {\bf rODENet-3}-N would be a practical choice in
terms of the parameter size, accuracy, stability, and execution time
\footnote{
	Performance improvement is still modest since some layers are executed by
	software. It would be further improved if weight parameters of more
	layers can be stored in BRAM. Although we used 32-bit fixed-point
	numbers, using reduced bit widths (e.g., 16-bit or less) can implement
	more layers in PL part.
}.

\section{Summary}\label{sec:conc}
To offload a part of ResBlock building blocks on PL part of low-cost
FPGA devices, in this paper we focused on ODENet and it was redesigned.
More specifically, as ODENet variants, reduced ODENets (rODENets) each
of which heavily uses a part of ODEBlocks and reduces some layers
differently were proposed and analyzed for low-cost FPGA devices.
We examined seven network architectures including the original ResNet
({\bf ResNet}-N), the original ODENet ({\bf ODENet}-N), and our
proposed rODENet variants (e.g., {\bf rODENet-3}-N) in terms of
the parameter size, accuracy, and execution time.
A part of ODEBlocks, such as layer1, layer2\_2, and layer3\_2, was
implemented on PL part of PYNQ-Z2 board to evaluate their FPGA
resource utilization.
For example, {\bf rODENet-3}-N heavily uses layer3\_2, reduces 
layer1, eliminates layer2\_2, and offloads layer3\_2 to PL part of the
FPGA.

The evaluation results using CIFAR-100 dataset showed that the
parameter sizes of {\bf rODENet-3}-N are 43.29\% and 81.80\% less than
those of {\bf ResNet}-N when N is 20 and 56, respectively.
The accuracies of {\bf rODENet-3}-N are the second highest next to
that of {\bf ResNet}-N when N is 20 and 32.
When N is 44 and 56, its accuracy is less than {\bf Hybrid-3}-N, but
it is still comparable to {\bf ODENet-3}-N.
{\bf rODENet-3}-N is 2.66 times faster than a pure software execution
when N is 56, which is the largest overall speedup by the FPGA.
It is 2.67 times faster than a software execution of 
{\bf ResNet}-N when N is 56.
In summary, our proposed {\bf rODENet-3}-N can strike a balance
between the parameter size, accuracy, stability, and execution time.

As a future work, we are working on the accuracy loss issue when the
adjoint method is used for training process.
Further experiments using more accurate ODE solvers, such as
Runge-Kutta method, are necessary.
Lastly, we are planning to offload the training process of the rODENet
variants to FPGA devices.


\end{document}